\documentclass{article}

\PassOptionsToPackage{numbers, compress}{natbib}
\usepackage[final]{nips_2017}

\usepackage[utf8]{inputenc} % allow utf-8 input
\usepackage[T1]{fontenc}    % use 8-bit T1 fonts
\usepackage{hyperref}       % hyperlinks
\usepackage{url}            % simple URL typesetting
\usepackage{booktabs}       % professional-quality tables
\usepackage{amsfonts}       % blackboard math symbols
\usepackage{nicefrac}       % compact symbols for 1/2, etc.
\usepackage{subcaption}
\usepackage{microtype}      % microtypography
\usepackage[table]{xcolor}
\usepackage{enumitem}
\usepackage[normalem]{ulem}

\hypersetup{
    colorlinks,
    linkcolor={red},
    citecolor={blue},
    urlcolor={blue}
}

\usepackage{tikz}
 \usepackage{forest}
\usetikzlibrary{shadows,arrows.meta}

\usepackage{algorithm}
\usepackage{algorithmic}
\usepackage{adjustbox,lipsum}
\usepackage{graphicx}   
\usepackage{tabularx}  
 \usepackage{mathtools}

\DeclarePairedDelimiter{\diagfences}{(}{)}
\newcommand{\diag}{\operatorname{diag}\diagfences}
\newcolumntype{C}[1]{>{\centering\let\newline\\\arraybackslash\hspace{0pt}}m{#1}}
\newcommand{\bmat}[1]{\left[\begin{matrix}
#1
\end{matrix}\right]}

\DeclareMathOperator*{\argmax}{arg\,max}

\newcolumntype{L}{>{\centering\arraybackslash}m{5.5cm}}

\newtheorem{teo}{Theorem}[section]

\newtheorem{lemma}{Lemma}[section]

\newtheorem{oss}{Remark}

	\def\R{\mathbb{R}}
	\def\P{\mathbb{P}}
	\def\E{\mathbb{E}}

\title{Hierarchical Methods of Moments}

\author{
  Matteo Ruffini
  \thanks{\small{mruffini@cs.upc.edu}} \\
  Universitat Politècnica \\de Catalunya \And
  Guillaume Rabusseau \thanks{\small{guillaume.rabusseau@mail.mcgill.ca}} \\
  McGill University\\
 \And
  Borja Balle \thanks{\small{pigem@amazon.co.uk}}\\
  Amazon Research \\Cambridge}

\begin{document}

\maketitle

\begin{abstract}
Spectral methods of moments provide a powerful tool for learning the parameters of latent variable models. Despite their theoretical appeal, the applicability of these methods to real data is still limited due to a lack of robustness to model misspecification. In this paper we present a hierarchical approach to methods of moments to circumvent such limitations. Our method is based on replacing the tensor decomposition step used in previous algorithms with approximate joint diagonalization. Experiments on topic modeling show that our method outperforms previous tensor decomposition methods in terms of speed and model quality.
\end{abstract}

\section{Introduction}

Unsupervised learning of latent variable models is a fundamental machine learning problem. Algorithms for learning a variety of latent variable models, including topic models, hidden Markov models, and mixture models are routinely used in practical applications for solving tasks ranging from representation learning to exploratory data analysis. For practitioners faced with the problem of training a latent variable model, the decades-old Expectation-Maximization (EM) algorithm~\cite{EMDempster} is still the tool of choice. Despite its theoretical limitations, EM owes its appeal to (i) the robustness of the maximum-likelihood principle to model misspecification, and (ii) the need, in most cases, to tune a single parameter: the dimension of the latent variables.

On the other hand, method of moments (MoM) algorithms for learning latent variable models via efficient tensor factorization algorithms have been proposed in the last few years~\cite{TensorLatent,SpectralLatent,SpectralLDA,jain2014learning,hsu2013learning, song2011spectral,EMSlow,chaganty2013spectral}. Compared to EM, moment-based algorithms provide a stronger theoretical foundation for learning latent variable models. In particular, it is known that in the realizable setting the output of a MoM algorithm will converge to the parameters of the true model as the amount of training data increases. Furthermore, MoM algorithms only make a single pass over the training data, are highly parallelizable, and always terminate in polynomial time. However, despite their apparent advantages over EM, the adoption of MoM algorithms in practical applications is still limited.

%\borja{Add more citations to MoMs. Anandkumar other papers, ICML'14 Balle et al., P. Liang group? Something by Eric Xing!}

Empirical studies indicate that initializing EM with the output of a MoM algorithm can improve the convergence speed of EM by several orders of magnitude, yielding a very efficient strategy to accurately learn latent variable models~\cite{EMSlow,chaganty2013spectral,bailly2011quadratic}. In the case of relatively simple models this approach can be backed by intricate theoretical analyses~\cite{zhang2014spectral}. Nonetheless, these strategies are not widely deployed in practice either.

%\borja{Add citations to papers combining MoMs and EM. ACML'11 Bailly, ICML'14 Balle et al., Mike Jordan NIPS'14?, etc.}

The main reason why MoM algorithms are not adopted by practitioners is their lack of robustness to model misspecification. Even when combined with EM, MoM algorithms fail to provide an initial estimate for the parameters of a model leading to fast convergence when the learning problem is too far from the realizable setting. For example, this happens when the number of the latent variables used in a MoM algorithm is too small to accurately represent the training data. In contrast, the model obtained by standalone EM in this case is reasonable and desirable: when asked for a small number of latent variables EM yields a model which is easy to interpret and can be useful for data visualization and exploration. For example, an important application of low-dimensional learning can be found in mixture models, where latent class assignments provided by a simple model can be used to split the training data into disjoint datasets to which EM is applied recursively to produce a hierarchical clustering \cite{steinbach2000comparison,savaresi2001performance}. The tree produced by such clusterings procedure provides a useful aid in data exploration and visualization even if the models learned at each branching point do not accurately represent the training data.

%\borja{Add citation to hierarchical k-means. Look at hierarchical clustering dropbox folder}

In this paper we develop a hierarchical method of moments that produces meaningful results even in misspecified settings. Our approach is different from previous attemps to design MoM algorithms for misspecified models. Instead of looking for convex relaxations of existing MoM algorithms like in \cite{DBLP:conf/icml/BalleQC12,DBLP:conf/nips/BalleM12,DBLP:conf/icml/QuattoniBCG14} or analyzing the behavior of a MoM algorithm with a misspecified number of latent states like in \cite{kulesza2014low,kulesza2015low}, we generalize well-known simultaneous diagonalization approaches to tensor decomposition by phrasing the problem as a non-convex optimization problem. Despite its non-convexity, the hierarchical nature of our method allows for a fast accurate solution based on low-dimensional grid search. We test our method on synthetic and real-world datasets on the topic modeling task, showcasing the advantages of our approach and obtaining meaningful results.
			
			%\borja{Mateo can provide a better description of the experiments}

%\borja{Maybe we can skip this related work, since it seems quite tangential.}

%\textbf{Related Work}.
%LOW-RANK SPECTRAL LEARNING. LOW-RANK TENSOR APPROXIMATIONS. MULTI-CLASS SUPERVISED LEARNING?

\section{Moments, Tensors, and Latent Variable Models}\label{sec:preliminaries}

%In this section we briefly recall how method of moments is used together with tensor decomposition techniques to learn LVMs parameters from data; we then analyze how state-of-the-art methods behave when the number of components is incorrectly specified, before outlining the goal of this paper.

%\borja{I'm starting to sharpen this up a bit...}

This section starts by recalling the basic ideas behind methods of moments for learning latent variable models via tensor decompositions. Then we review existing tensor decomposition algorithms and discuss the effect of model misspecification on the output of such algorithms.

For simplicity we consider first a \emph{single topic model} with $k$ topics over a vocabulary with $d$ words. A single topic model defines a generative process for text documents where first a topic $Y \in [k]$ is drawn from some discrete distribution $\P[Y = i] = \omega_i$, and then each word $X_t \in [d]$, $1 \leq t \leq T$, in a document of length $T$ is independently drawn from some distribution $\P[X_t = j | Y = i] = \mu_{i,j}$ over words conditioned on the document topic. The model is completely specified by the vector of topic proportions $\omega \in \R^k$ and the word distributions $\mu_i \in \R^d$ for each topic $i \in [k]$. We collect the word distributions of the model as the columns of a matrix $M = [\mu_1 \cdots \mu_k] \in \R^{d \times k}$.

It is convenient to represent the words in a document using one-hot encodings so that $X_t \in \R^d$ is an indicator vector. With this notation, the conditional expectation of any word in a document drawn from topic $i$ is $\E[X_t | Y = i] = \mu_i$, and the random vector $X = \sum_{t=1}^T X_t$ is conditionally distributed as a multinomial random variable, with parameters $\mu_i$ and $T$. Integrating over topics drawn from $\omega$ we obtain the first moment of the distribution over words $M_1 = \E[X_t] = \sum_{i} \omega_i \mu_i = M \omega$. Generalizing this argument to pairs and triples of distinct words in a document yields the matrix of second order moments and the tensor of third order moments of a single topic model:
\begin{align}
M_2 &= \E[X_{s} \otimes X_{t}] = \sum_{i} \omega_i \mu_i \otimes \mu_i \in \R^{d \times d} \enspace,\label{eqn:mom2}  \\
M_3 &= \E[X_{r} \otimes X_{s} \otimes X_{t}] = \sum_{i} \omega_i \mu_i \otimes \mu_i \otimes \mu_i \in \R^{d \times d \times d} \enspace, \label{eqn:mom3}
\end{align}
where $\otimes$ denotes the tensor (Kronecker) product between vectors. By defining the matrix $\Omega = \diag{\omega}$ one also obtains the expression $M_2 = M \Omega M^\top$.  

A method of moments for learning single topic models proceeds by (i) using a collection of $n$ documents to compute empirical estimates $\hat{M}_1$, $\hat{M}_2$, $\hat{M}_3$ of the moments, and (ii) using matrix and tensor decomposition methods to (approximately) factor these empirical moments and extract the model parameters from their decompositions. From the algorithmic point of view, the appeal of this scheme resides in the fact that step (i) requires a single pass over the data which can be trivially parallelized using map-reduce primitives, while step (ii) only requires linear algebra operations whose running time is independent of $n$. The specifics of step (ii) will be discussed in Section~\ref{sec:prevtensor}.

Estimating moments $\hat{M}_{m}$ from data with the property that $\E[\hat{M}_m] = M_m$ for $m \in \{1,2,3\}$ is the essential requirement for step (i). In the case of single topic models, and more generally multi-view models, such estimations are straightforward. For example, a simple consistent estimator takes a collection of documents $\{x^{(i)}\}_{i=1}^n$ and computes $\hat{M}_3 = (1/n) \sum_{i=1}^n x^{(i)}_1 \otimes x^{(i)}_2 \otimes x^{(i)}_3$ using the first three words from each document. More data-efficient estimators for datasets containing long documents can be found in the literature \cite{ruffini2016new}.

%for now I just cite them, if we have time I'll add them
%\borja{We can provide Mateo's (unpublished?) estimates in the supplementary}

For more complex models the method sketched above requires some modifications. Specifically, it is often necessary to correct the statistics directly observable from data in order to obtain vectors/matrices/tensors whose expectation over a training dataset exhibits precisely the relation with the parameters $\omega$ and $M$ described above. For example, this is the case for Latent Dirichlet Allocation and mixtures of spherical Gaussians \cite{SpectralLDA, hsu2013learning}. For models with temporal dependence between observations, e.g.\ hidden Markov models, the method requires a spectral projection of observables to obtain moments behaving in a multi-view-like fashion \cite{SpectralLatent,SpectralLatentHMM}. Nonetheless, methods of moments for these models and many others always reduces to the factorization of a matrix and tensor of the form $M_2$ and $M_3$ given above.

%\borja{Maybe we don't need so many details in this last paragraph...}

\subsection{Existing Tensor Decomposition Algorithms}\label{sec:prevtensor}

Mathematically speaking, methods of moments attempt to solve the polynomial equations in $\omega$ and $M$ arising from plugging the empirical estimates $\hat{M}_m$ into the expressions for their expectations given above. Several approaches have been proposed to solve these non-linear systems of equations.

A popular method for tensor decomposition is Alternating Least Squares (ALS) \cite{kolda2009tensor}. Starting from a random initialization of the factors composing a tensor, ALS iteratively fixes two of the three factors, and updates the remaining one by solving an overdetermined linear least squares problem. ALS is easy to implement and to understand, but is known to be prone to local minima, needing several random restarts to  yield meaningful results. These limitations fostered the research for methods with  guarantees, that, in the unperturbed setting, optimally decompose a tensor like the one in Eq.~\eqref{eqn:mom3}. We now briefly analyze some of these methods.

The tensor power method (TPM)  \cite{TensorLatent} starts with a whitening step where, given the SVD $M_2=U S U^\top$, the whitening matrix $E=U S^{1/2} \in \R^{d \times k}$ is used to transform $M_3$ into a symmetric orthogonally decomposable tensor 
\begin{equation}\label{eq:TPMT}
T=\sum_{i=1}^k \omega_i E^\dagger\mu_i \otimes E^\dagger\mu_i \otimes E^\dagger\mu_i \in \R^{k\times k \times k}
\end{equation}
The weights $\omega_i$ and vectors $\mu_i$ are then recovered from $T$ using a tensor power method and inverting the whitening step.

The same whitening matrix is used in \cite{SpectralLatent,SpectralLDA}, where the authors observe that the whitened slices of $M_3$ are simultaneously diagonalized by the Moore-Penrose pseudoinverse of $M \Omega^{1/2}$. Indeed, since $M_2 = M \Omega  M^\top = E E^\top$, there exists a unique orthonormal matrix $O \in \R^{k \times k}$ such that $M \Omega^{1/2} = E O$. Writing $M_{3,r} \in \R^{d \times d}$ for the $r$th slice of $M_3$ across its second mode and $m_r$ for the $r$th row of $M$, it follows that
\begin{equation*}
M_{3,r} =  M \Omega^{1/2}  diag(m_r) \Omega^{1/2} M^\top = E O diag(m_r) O^\top E^\top \enspace.
\end{equation*}
Thus, the problem can be reduced to searching for the common diagonalizer $O$ of the whitened slices of $M_3$ defined as
\begin{equation}\label{eq:hr}
H_{r} = E^{\dagger} M_{3,r} E^{\dagger\top} = O diag(m_r) O^\top \enspace.
\end{equation}
In the noiseless settings it is sufficient to diagonalize any of the slices $M_{3,r}$. However, one can also recover $O$ as the eigenvectors of a random linear combination of the various $H_r$ which is more robust to noise \cite{SpectralLatent}.

Lastly, the method proposed in~\cite{kuleshov2015tensor} consists in directly performing simultaneous diagonalization of random linear combinations of slices of $M_3$ without any whitening step. This method, which in practice is slower than the others (see Section \ref{sec:synth}), under an incoherence assumption on the vectors $\mu_i$, can robustly recover the weights $\omega_i$ and vectors $\mu_i$ from the tensor $M_3$, even when it is not orthogonally decomposable.

\subsection{The Misspecified Setting}\label{subsec:misspecified}

The methods listed in the previous section have been analyzed in the case where the algorithm only has access to noisy estimates of the moments. However, such analyses assume that the data was generated by a model from the hypothesis class, that the matrix $M$ has rank $k$, and that this rank is known to the algorithm. In practice the dimension $k$ of the latent variable can be cross-validated, but in many cases this is not enough: data may come from a model outside the class, or from a model with a very large true $k$. Besides, the moment estimates might be too noisy to provide reliabe estimates for large number of latent variables. It is thus frequent to use these algorithms to estimate $l < k$ latent variables. However, existing algorithms are not robust in this setting, as they have not been designed to work in this regime, and there is no theoretical explanation of what their outputs will be. 

%This is the starting hypothesis of the even the ``best'' $k$ yields poor models due to a mismatch between the process that generated the data and the model the algorithm is trying to learn. have no paper. What happens is that (i) whole paper, so it needs to be very compellin In practice, when working on real  world data, the number $k$ can only be inferred; it is thus likely to use these methods in the misspecified setting where a wrong rank $l < k$ is plug in the algorithm. 
% 
%The simultaneous diagonalization method based on optimization %in principle may produce a matrix that nearly diagonalizes the %slices of $M_3$, but in \cite{kuleshov2015tensor} is not %provided an analysis of what happens when running the method %asking for $l<k$, nor on who are going to be the results.

The methods relying on a whitening step~\cite{TensorLatent,SpectralLatent,SpectralLDA}, will perform the whitening using the matrix $ E_l^{\dagger}$ obtained from the low-rank SVD truncated at rank $l$:  $M_2 \approx U_lS_lU_l^\top = E_lE_l^\top$. TPM will use $E_l$ to whiten the tensor $M_3$ to a tensor $T_l\in\R^{l\times l \times l}$.  However, when $k>l$, $T_l$ may not admit a symmetric orthogonal decomposition \footnote{See the supplementary material for an example corroborating this statement.}. Consequently, it is not clear what  TPM will return in this case and there are no guarantees it will even converge.
%
%The Methods from \cite{SpectralLatent,SpectralLDA,TensorLatent}, that use the whitening process, will perform the whitening using the matrix $ E_l^{\dagger}$, obtained from the low-rank SVD truncated at the $l$-th eigenvector:  $M_2 \approx U_lS_lU_l^\top = E_lE_l^\top$. TPM, will use $E_l$ to whiten the tensor $M_3$ to a tensor $T_l\in\R^{l\times l \times l}$.  However, when $k>l$ there are no guarantees that $T_l$ will be close to admitting a symmetric orthogonal decomposition decomposition. Consequently, it is not clear what the iterative power method will return in this case; also there are no guarantees that TPM applied to $T_l$ will even converge. 
%
The methods from \cite{SpectralLatent,SpectralLDA} will compute the matrices $ H_{l,r} = E_l^{\dagger} M_{3,r}E_l^{\dagger\top}$ for $r \in [d]$ that may  not be jointly diagonalizable, and in this case there is no theoretical justification of what what the result of these algorithms will be.
Similarly, the simultaneous diagonalization method proposed in~\cite{kuleshov2015tensor} produces a matrix that nearly diagonalizes the slices of $M_3$, but no analysis is given for this setting.

\section{Simultaneous Diagonalization Based on Whitening and Optimization}
This section presents the main contribution of the paper: a simultaneous diagonalization algorithms based on whitening and optimization we call SIDIWO (\textbf{Si}multaneous \textbf{Di}agonalization based on \textbf{W}hitening and \textbf{O}ptimization). When asked to produce $l = k$ components in the noiseless setting, SIDIWO will return the same output as any of the methods discussed in Section~\ref{sec:prevtensor}. However, in contrast with those methods, SIDIWO will provide useful results with a clear interpretation even in a misspecified setting ($l < k$).

\subsection{SIDIWO in the Realizable Setting}\label{sec:optstpt}
To derive our SIDIWO algorithm we first observe that in the noiseless setting and when $l = k$, the pair $(M,\omega)$ returned by all methods described in Section~\ref{sec:prevtensor} is the solution of the optimization problem given in the following lemma\footnote{The proofs of all the results are provided in the supplementary material.}.
% To derive our SIDIWO algorithm we observe that in the ideal setting (no noise, $l = k$) the parameters $(M,\omega)$ produced as solution by all methods described in Section~\ref{sec:prevtensor} can also be obtained through the solution of an optimization problem \footnote{The proofs are provided in the supplementary material.}.
\begin{lemma}\label{lemma:1}
Let $M_{3,r}$ be the $r$-th slice across the second mode of the tensor $M_3$ from \eqref{eqn:mom3} with parameters $(M,\omega)$. Suppose $\mathop{rank}(M) = k$ and let $\Omega = \diag{\omega}$. Then the matrix $(M\Omega^{1/2})^\dagger$ is the unique optimum (up to column rescaling) of the optimization problem
\begin{equation}\label{eq:constropt}
\min_{D\in \mathcal{D}_k} \sum_{i\neq j}\left(\sum_{r=1}^d (D M_{3,r}D^\top)^2_{i,j} \right)^{1/2} \enspace,
\end{equation}
where $\mathcal{D}_k = \{D: D = (E O_k)^\dagger \text{ for some } O_k \text{ s.t. } O_kO_k^\top = \mathbb{I}_k\}$ and $E$ is the whitening matrix defined in Section~\ref{sec:prevtensor}.
\end{lemma}

\begin{oss}[The role of the constraint]\label{oss:dk}
Consider the cost function of Problem \eqref{eq:constropt}: in an unconstrained setting, there may be several matrices minimizing that cost. A trivial example is the zero matrix. A less trivial example is when the rows of $D$ belong to the orthogonal complement of the column space of the matrix $M$. The constraint $D = (E O_k)^\dagger$ for some orthonormal matrix $O_k$ first excludes the zero matrix from  the set of feasible solutions, and second guarantees that all feasible solutions lay in the space generated by the columns of $M$. 
\end{oss}

%\begin{oss}[The role of the constraint]\label{oss:dk}
%Consider the cost function of Problem \eqref{eq:constropt}: in an unconstrained setting, there may be several matrices minimizing that cost. A trivial example is the matrix $D$ with all entries at zero. A less trivial case may be when the rows of $D$ belong to the orthogonal complement in $\R^n$ of the space generated by columns of the matrix $M$. The constraint of requiring $D = (E_kO_k)^\dagger$ for some orthonormal matrix, first excludes the zero matrix from set of feasible solutions, and second guarantees that all the feasible solutions lay in the same space generated by the columns of $M$. 
%\end{oss}
 
\begin{algorithm}[h!]
\caption{SIDIWO: {Si}multaneous {Di}agonalization based on {W}hitening and {O}ptimization}
\label{alg:1}
\begin{algorithmic}[1]
\REQUIRE $M_1$, $M_2$, $M_3$, the number of latent states $l$ 
\STATE Compute a SVD of $M_2$ truncated at the $l$-th singular vector: $M_2 \approx U_lS_lU_l^\top$.
\STATE Define the matrix $E_l =U_lS_l^{1/2}\in \R^{d\times l}$.
\STATE Find the matrix $D\in \mathcal{D}_l$ optimizing Problem \eqref{eq:constropt}.
\STATE Find  $(\tilde{M},\tilde{\omega})$ solving
$
{\begin{cases} 
 \tilde{M} \tilde{\Omega}^{1/2} = D^\dagger\\
 \tilde{M}\tilde{\omega}^\top = M_1
 \end{cases}}
$
\RETURN $(\tilde{M},\tilde{\omega})$

\end{algorithmic}
\end{algorithm}
Problem \eqref{eq:constropt} opens a new perspective on using simultaneous diagonalization to learn the parameters of a latent variable model. In fact, one could recover the pair $(M,\omega)$ from the relation $M\Omega^{1/2} = D^\dagger$ by first finding the optimal $D$ and then individually retrieving $M$ and $\omega$ by solving a linear system using the vector $M_1$. This approach, outlined in Algorithm \ref{alg:1}, is an alternative to the ones presented in the literature up to now (even though in the noiseless, realizable setting, it will provide the same results). Similarly to existing methods, this approach requires to know the number of latent states. We will however see in the next section that Algorithm~\ref{alg:1} provides meaningful results even when a misspecified number of latent states $l<k$ is used.
%\borja{Does the optimal $O_k$ play a role in all this? Is it useful to implement line 4 in Algorithm 1?}

%Problem \eqref{eq:constropt} opens a new perspective on simultaneous diagonalization to recover the parameters of a LVM. In fact, one could find the optimal $D$ to  recover the pair $(M,\omega)$ from $M\Omega^{1/2} = D^\dagger$. $M$ and $\omega$ are then retrieved individually using the vector $M_1$, solving a linear system. This approach, outlined in Algorithm~\ref{alg:1}, is alternative to the ones presented in the literature up to now, even if in the noiseless, well-specified setting, will provide the same results. Like the existing methods, this approach requires to know the number of latent states: the whitening matrix in fact is retrieved from the first $k$ singular vectors of $M_2$; however we will see in the next section that \ref{alg:1} provides meaningful results even when inputted with misspecified number of latest states $l<k$.

\subsection{The Misspecified Setting}
Algorithm~\ref{alg:1} requires as inputs the low order moments  $M_1$, $M_2$, $M_3$ along with the desired number of latent states $l$ to recover. If $l=k$, it will return the exact model parameters $(M,\omega)$; we will now see that it will also provide meaningful results when $l<k$.  In this setting, Algorithm~\ref{alg:1} returns a pair $(\tilde{M},\tilde{\omega})\in \R^{d\times l}\times \R^l$ such that the matrix $D = (\tilde{M} \tilde{\Omega}^{1/2})^\dagger$ is optimal for the optimization problem
\begin{equation}\label{eq:constroptl}
\min_{D\in \mathcal{D}_l} \sum_{i\neq j}\left(\sum_{r=1}^d (D M_{3,r}D^\top)^2_{i,j}\right)^{1/2}.
\end{equation}   
{ 
Analyzing the space of feasible solutions (Theorem \ref{prop:whiten}) and the optimization function (Theorem \ref{teo:newopt}), we will obtain theoretical guarantees on what SIDIWO returns when $l<k$, showing that the trivial solutions are not feasible, and that, in the space of feasible solutions, SIDIWO’s optima will approximate the true model parameters according to an intuitive geometric interpretation. }
\paragraph{Remarks on the constraints.}
The first step consists in analyzing the space of feasible solutions $ \mathcal{D}_l$ when $l<k$. The observations outlined in Remark \ref{oss:dk} still hold in this setting: the zero solution and the matrices laying in the orthonormal complement of $M$ are not feasible.  
Furthermore, the following theorem shows that other undesirable solutions will be avoided. 

%\paragraph{Remarks on the constraints.}
%The first step to understand $(\tilde{M},\tilde{\omega})$, is to analyze the space of feasible solutions $ \mathcal{D}_l$ when $l<k$. The observations outlined in Remark \ref{oss:dk}, are still valid here: the zero solution, and the matrices laying the orthonormal complement of $M$ will not be feasible.  
%Furthermore, some other undesirable results will be cut out from the games in this setting, as the following Theorem shows. 

\begin{teo}\label{prop:whiten}
Let $D\in \mathcal{D}_l$ with rows $d_1,...,d_l$, and let $\mathbb{I}_{r,s}$ denote the $r\times s$ identity matrix.  The following facts hold under the hypotheses of Lemma \ref{lemma:1}:
\begin{enumerate}[itemsep=-1pt, topsep=-1pt, partopsep=0pt]
\item  For any row $d_i$, there exists at least one column of $M$ such that  $\langle d_i,\mu_j \rangle \neq 0$.
\item The columns of any  $\tilde{M}$ satisfying $ {\tilde{M}} {\tilde{\Omega}}^{1/2} = D^\dagger$  are a linear combination of those of $M$, laying in the best-fit $l$-dimensional subspace of the space spanned by the columns of $M$.
\item Let $\pi$ be any permutation of $\{1,...,d\}$, and let $M_\pi$ and $\Omega_\pi$ be obtained by permuting the columns of $M$ and $\Omega$ according to $\pi$. If $\langle \mu_i,\mu_j\rangle  \neq 0$ for any $i,j$, then $((M_\pi\Omega_\pi^{1/2})\mathbb{I}_{k,l})^\dagger \notin \mathcal{D}_l $, and similarly $ \mathbb{I}_{l,k}(M_\pi\Omega_\pi^{1/2})^\dagger\notin \mathcal{D}_l$.
\end{enumerate}
\end{teo}

%\begin{teo}\label{prop:whiten}
%Let $D\in \mathcal{D}_l$ with rows $d_1,...,d_l$. The following facts hold:
%\begin{enumerate}
%\item  For any row $d_i$, there exists at least one column of $M$ such that  $\langle d_i,\mu_j \rangle \neq 0$.
%\item The columns of any  $\tilde{M}$, satisfying $ {\tilde{M}} {\tilde{\Omega}}^1/2 = D^\dagger$,  are a linear combination of those of $M$, laying in the best-fit $l$-dimensional subspace of the space spanned by the columns of $M$.
%\item Let $\pi$ be any permutation of $\{1,...,n\}$, and let $M_\pi$ and $\Omega_\pi$ be $M$ and $\Omega$, with columns permuted according to $\pi$. If $\langle \mu_i,\mu_j\rangle  \neq 0$ for any $i,j$, then $((M_\pi\Omega_\pi^{1/2})\mathbb{I}_{k,l})^\dagger \notin \mathcal{D}_l $, and similarly $ \mathbb{I}_{l,k}(M_\pi\Omega_\pi^{1/2})^\dagger\notin \mathcal{D}_l$.
%\end{enumerate}

%\end{teo}
The second point of Theorem \ref{prop:whiten} states that the feasible solutions will lay in the best $l$-dimensional subspace approximating the one spanned by the columns of $M$. This has two interesting consequences: if the columns of $M$ are not orthogonal, point 3 guarantees that  $\tilde{M}$ cannot simply be a sub-block of the original $M$, but rather a non-trivial linear combination of its columns laying in the best $l$-dimensional subspace approximating its column space. In the single topic model case with $k$ topics, when asked to recover $l<k$ topics, Algorithm~\ref{alg:1} will not return a subset of the original $k$ topics, but a matrix $\tilde{M}$ whose columns gather the original topics via 
a non trivial linear combination: the original topics will all be represented in the columns of $\tilde{M}$ with different weights.
When the columns of $M$ are orthogonal, this space coincides with the space of the  $l$ columns of $M$ associated with the $l$ largest $\omega_i$; in this setting, the matrix  $(M_\pi\Omega_\pi^{1/2})\mathbb{I}_{k,l}$~(for some permutation $\pi$) is a feasible solution and minimizes Problem~\eqref{eq:constroptl}. Thus, Algorithm~\ref{alg:1} will recover the top $l$ topics.

\paragraph{Interpreting the optima.} 
Let $\tilde{M}$ be such that $D = (\tilde{M}\tilde{\Omega}^{1/2})^\dagger \in \mathcal{D}_l$ is a minimizer of Problem~\eqref{eq:constroptl}. In order to better understand the relation between $\tilde{M}$ and the original matrix $M$, we will show that the cost function of Problem~\eqref{eq:constroptl} can be written in an equivalent form, that unveils a geometric interpretation. 
%Assume to have a matrix $\tilde{M}$, such that $D = (\tilde{M}\tilde{\Omega}^{1/2})^\dagger \in \mathcal{D}_l$ is optimal for Problem \eqref{eq:constroptl}. We want to understand who this matrix is with respect to the original matrix $M$. To perform this, we will focus on the cost function of the Problem \eqref{eq:constroptl}, showing that it can be written in an equivalent form, that unveils a geometric interpretation. 
 
\begin{teo}\label{teo:newopt}
Let $d_1,...,d_l$ denote the rows of $D\in \mathcal{D}_l$  and introduce the following optimization problem %\eqref{eq:constropt-center}
\begin{equation}\label{eq:constropt-center}
\min_{D\in\mathcal{D}_l}  \sum_{i\neq j} \sup_{v\in \mathcal{V}_M} \sum_{h = 1}^{k}\langle d_i,\mu_h\rangle\langle d_j,\mu_h\rangle\omega_h v_h 
\end{equation}
where
$
\mathcal{V}_M = \{v\in\R^k: v = \alpha^\top M,  \text{where } \|\alpha\|_2 \leq 1\}
$.
Then this problem is equivalent to \eqref{eq:constroptl}.
\end{teo}
%\borja{Are you using the assumptions that $M$ is full rank in any of these results? This should be mentioned in the statements.}
First, observe that the cost function in Equation~\eqref{eq:constropt-center} prefers $D$'s such that the vectors 
$u_i =[\langle d_i,\mu_1\sqrt{\omega_1}\rangle ,...,\langle d_i,\mu_k\sqrt{\omega_k}\rangle]$, $i \in [l]$, have disjoint support. This is a consequence of the $\sup_{v \in \mathcal{V}_M}$, and requires that, for each $j$, the entries $\langle d_i,\mu_j\sqrt{\omega_j}\rangle$ are close zero for at least all but one of the various $d_i$. Consequently, each center will be almost orthogonal to all but one row of the optimal $D$; however the number of centers is greater than the number of rows of $D$, so the same row $d_i$ may be nonorthogonal to various centers.

%The cost function in Equation~\eqref{eq:constropt-center} is  trying to maximize the disjoint support of the various $v_i$. Indeed, the $j$-th entry of $v_i$ is zero if  $d_i$ is orthogonal to $\mu_j\sqrt{\omega_j}$;
%however the trivial solutions of this problem are removed from the constraint; also the solutions that samples $l$ columns of $M\Omega^{1/2}$ and get $D$ via pseudoinversion, are forbidden, as well as those that take $l$ rows from $(M\Omega^{1/2})^\dagger$, that in principle minimize the cost.
%however, by the constraint, $d_i$ can not be orthogonal to all the vectors $\mu_j$, so, for any vector $\mu_j$ there will be for sure a $d_i$ such that $<d_i,\mu_1\sqrt{\omega_1}>\neq 0$.

For illustration, consider the single topic model: a solution $D$ to Problem \eqref{eq:constropt-center}  would have rows that should be  as orthogonal as possible to some topics and as aligned as possible to the others; in other words, for a given topic $j$, the optimization problem is trying to set  $\langle d_i,\mu_j\sqrt{\omega_j} \rangle = 0$ for all but one of the various $d_i$. Consequently, each column of the output $\tilde{M}$ of Algorithm~\ref{alg:1} should be in essence aligned with some of the topics and orthogonal to the others.

It is worth mentioning that the constraint set $\mathcal{D}_l$ forbids the trivial solutions such as the zero matrix, the pseudo-inverse of any subset of $l$ columns of $M\Omega^{1/2}$, and any subset of $l$ rows of $(M\Omega^{1/2})^\dagger$~(which all have an objective value of $0$). 

We remark that Theorem \ref{teo:newopt} doesn't  require the matrix $M$ to be full rank $k$: we only need it to have at least rank greater or equal to $l$, in order to guarantee that the constraint set $\mathcal{D}_l$ is well defined.

%Consider the example of the single topic model, where the $\mu_j$ are the $k$ centers of the mixture: minimizing Problem \eqref{eq:constropt-center}, we are asking for a solution $D$ whose rows should be the most orthogonal as possible to some topics and the most aligned as possible to some others; in other words, for a given topic $j$, the optimization problem is trying to set  $\langle d_i,\mu_j\sqrt{\omega_j} \rangle = 0$ for all but-one the various $d_i$. In this sense, if we then take the outputs of Algorithm~\ref{alg:1}, $(\tilde{M},\tilde{\omega})$,  we will have each of the columns of $\tilde{M}$ that try to be aligned with some of the topics and orthogonal to some others. 

 %This method will then be used to perform hierarchical topic modeling in~Section \ref{sec:lda}.

\paragraph{An optimal solution when $l=2$.}
While Problem~\eqref{eq:constropt} can be solved in general using an extension of the Jacobi technique~\cite{cardoso1996jacobi,bunse1993numerical}, we provide a simple and efficient method for the case $l=2$. This method will then be used to perform hierarchical topic modeling in~Section \ref{sec:lda}.
When $l = 2$, Equation~\eqref{eq:constroptl} can be solved optimally with few simple steps; in fact, the following theorem shows that solving~\eqref{eq:constroptl} is equivalent to minimizing a continuous function on the compact one-dimensional set $I = [-1,1]$, which can easily be done by griding $I$. Using this in Step 3 of Algorithm~\ref{alg:1}, one can efficiently compute an arbitrarily good approximation of the optimal matrix $D\in \mathcal{D}_2$.

\begin{teo}\label{teo:l2}
Consider the continuous function
$
F(x) = c_1 x^4 + c_2 x^3\sqrt{1-x^2} + c_3 x\sqrt{1-x^2} + c_4  x^2 + c_5
$, where the coefficients $c_1,...,c_5$ are functions of the entries of $M_2$ and $M_3$. Let $a$ be the minimizer of $F$
on $[-1,1]$, and consider the matrix 
\begin{equation*} 
O_{a} = 
  \begin{bmatrix}
    \sqrt{1-a^2} &a\\
    -a & \sqrt{1-a^2}
  \end{bmatrix} \enspace.
\end{equation*} 
Then, the matrix $D = (E_2 O_a)^\dagger$ is a minimizer of  Problem \eqref{eq:constroptl} when $l=2$.
\end{teo}
\section{Case Study: Hierarchical Topic Modeling}\label{sec:lda}
In this section, we show how SIDIWO can be used to efficiently recover hierarchical representations of latent variable models. Given a latent variable model with $k$ states, our method allows to recover a pair $(\tilde{M},\tilde{\omega})$ from estimate of the moments $M_1$, $M_2$ and $M_3$, where the $l$ columns of $\tilde{M}$  offer a synthetic representation of the $k$ original centers. We will refer to these $l$ vectors as \emph{pseudo-centers}: each  pseudo-center is representative of a group of the original centers. Consider the case $l=2$. A dataset  $\mathcal{C}$ of $n$ samples can be split into  two smaller subsets according to their similarity to the two pseudo-centers. Formally, this assignment is done using Maximum A Posteriori (MAP) to find the pseudo-center giving maximum conditional likelihood to each sample. The splitting procedure can be iterated recursively to obtain a divisive binary tree, leading to a hierarchical clustering algorithm. { While this hierarchical clustering method can be applied to any latent variable model that can be learned with the tensor method of moments (e.g. Latent Dirichlet Allocation), we present it here for the single topic model for the sake of simplicity.} 

% {\color{red} We present now a specialization to the single topic model; extensions to more complex models like LDA are possible, and can be performed following  the same methodology presented in this section.}
We consider a corpus  $\mathcal{C}$ of $n$ texts encoded as in Section \ref{sec:preliminaries} and we split $\mathcal{C}$ into  two smaller corpora according to their similarity to the two pseudo-centers in two steps: project the pseudo-centers on the simplex to obtain discrete probability distributions~(using for example the method described in~\cite{duchi2008efficient}), and use MAP assignment to assign each text $x$ to a pseudo-center. This process is summarized in
Algorithm~\ref{alg:2}.
\begin{algorithm}[h!]
\caption{Splitting a corpus into two parts}
\label{alg:2}
\begin{algorithmic}[1]
\REQUIRE A corpus of texts $\mathcal{C} = (x^{(1)},...,x^{(n)})$.
\STATE Estimate $M_1$, $M_2$ and $M_3$.
\STATE Recover $l=2$ pseudo-center with Algorithm~\ref{alg:1} .
\STATE Project the Pseudo-center to the simplex
\FOR{$i \in [n]$}
\STATE Assign the text $x^{(i)}$  to the cluster $\mathop{Cluster}(i) = \argmax_{j} \,\P[X = x^{(i)} | Y = j, \tilde{\omega}, \tilde{M}]$,  where $\P[X  | Y = j, \tilde{\omega}, \tilde{M}]$
is the multinomial distr. associated to the $j$-th projected pseudo-center.
\ENDFOR
\RETURN The cluster assignments $\mathop{Cluster}$.
\end{algorithmic}
\end{algorithm}
Once the corpus $\mathcal{C}$ has been split into two subsets $\mathcal{C}_1$ and $\mathcal{C}_2$, each of these subsets may still contain the full set of topics but the topic distribution will differ in the two: topics similar to the first pseudo-center will be predominant in the first subset, the others in the second. By recursively iterating this process, we obtain a binary tree where topic distributions in the nodes with higher depth are expected to be more concentrated on fewer topics. 

In the next sections, we assess the validity of this approach on both synthetic and real-world data\footnote{The experiments in this section have been performed in Python 2.7, using \textit{numpy} \citep{van2011numpy} library for linear algebra operations, with the exception of the implementation of the method from \cite{kuleshov2015tensor}, for which we used the author's Matlab implementation: \url{https://github.com/kuleshov/tensor-factorization}.  All the experiments were run on a MacBook Pro with an Intel Core i5 processor. The implementation of the described algorithms can be found a this link: \url{https://github.com/mruffini/Hierarchical-Methods-of-Moments}.}.
% In this way, we can split $\mathcal{C}$ into two subsets: $\mathcal{C}_1$ and $\mathcal{C}_2$. The idea of this splitting, is that, even if  $\mathcal{C}_1$ and $\mathcal{C}_2$ may still contain the full set of topics, the topic distribution will differ in the two subsets: the topics similar to the first pseudo-topic will be predominant in the first subset, the others in the second. 

% The procedure can be iterated many times recursively: we can split $\mathcal{C}_1$ into two subsets and then split again the resulting subsets, creating a divisive binary tree, calling recursively Algorithm~\ref{alg:2}; this will provide a divisive hierarchical clustering algorithm for the single topic model, easily generalizable to any mixture model.

%\subsection{Experiments on Synthetic Data}
% We now experimentally test the proposed method on synthetic and real-world data\footnote{The experiments in this section have been performed in Python 2.7, using \textit{numpy} \citep{van2011numpy} library for linear algebra operations, with the exception of the implementation of the method from \cite{kuleshov2015tensor}, for which we used the author's matlab implementation at this link: \url{https://github.com/kuleshov/tensor-factorization}.  All the experiments have run on a MacBook Pro, with an Intel Core i5 processor.}.

\subsection{Experiment on Synthetic Data}\label{sec:synth}

\begin{figure}\label{fig:Topic}
    \centering
    \begin{subfigure}[b]{0.27\textwidth}
            \centering
            \includegraphics[width=\textwidth]{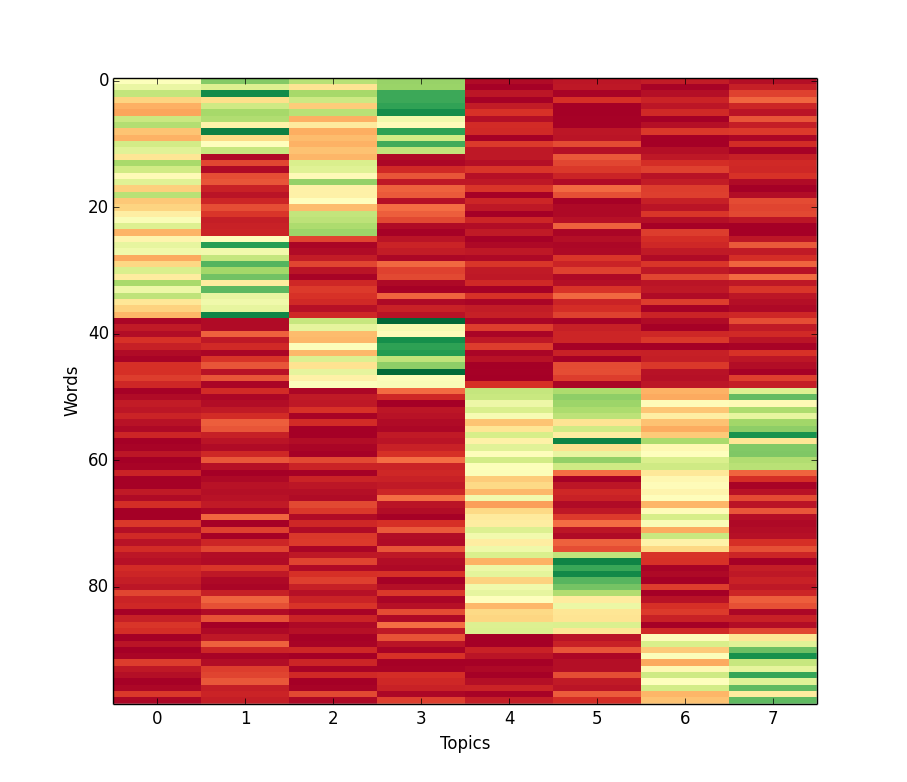}
            \caption{}
            \label{fig:TheTopic}
    \end{subfigure}
        \centering
    \begin{subfigure}[b]{0.31\textwidth}
            \centering
            \includegraphics[width=\textwidth]{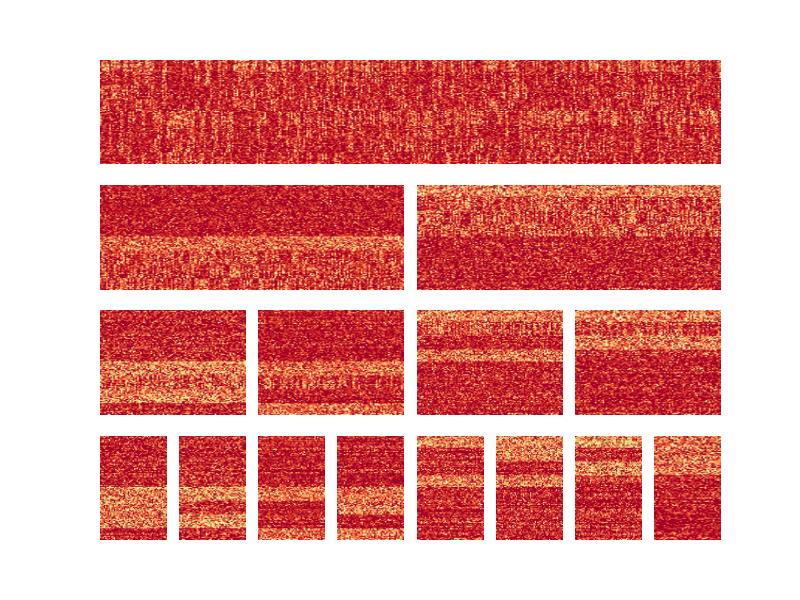}
            \caption{}
            \label{fig:Cheatmap}
    \end{subfigure}
\begin{subfigure}[b]{0.33\textwidth}
            \centering
            \includegraphics[width=\textwidth]{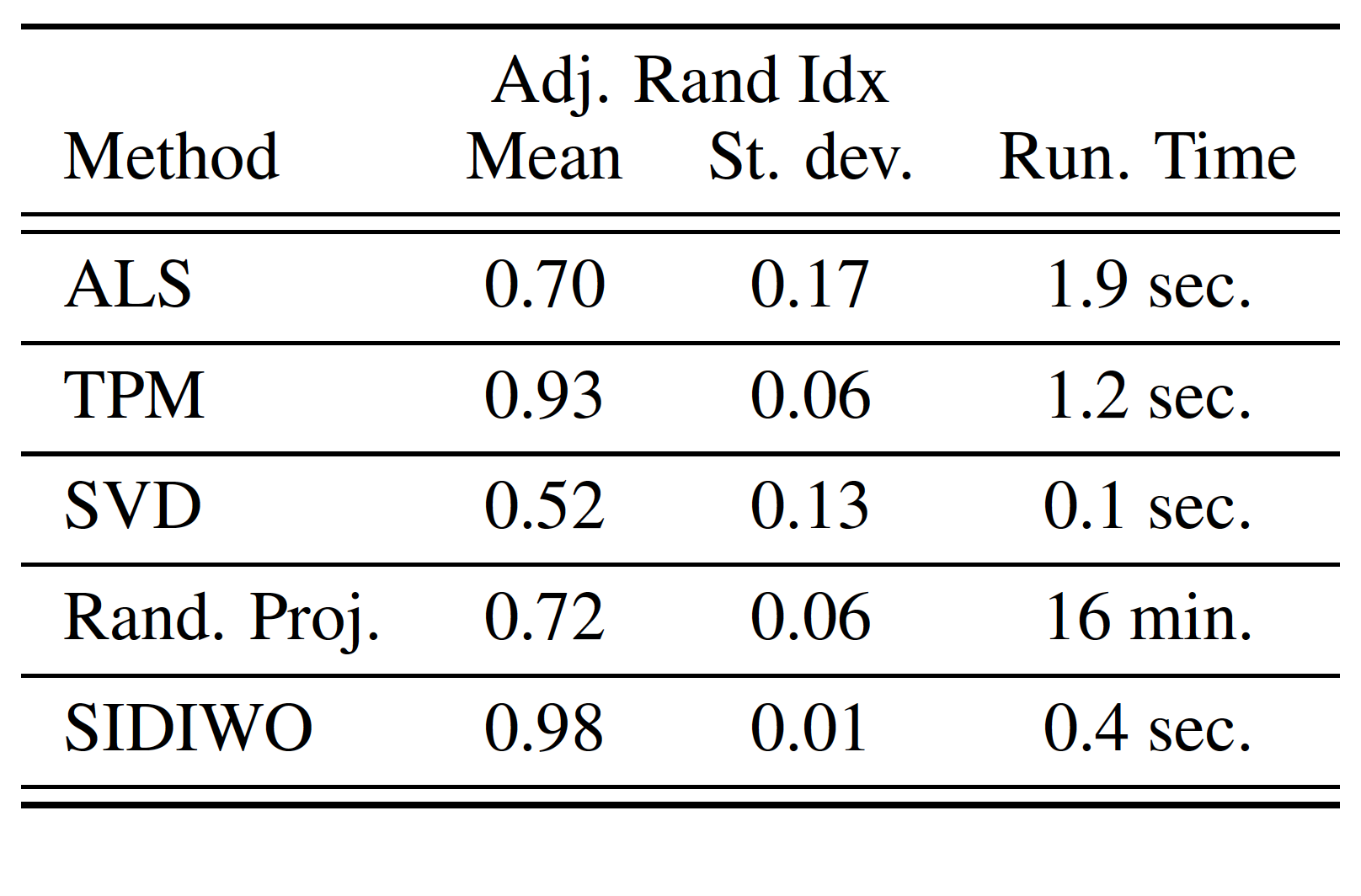}
            \caption{}
            \label{fig:tab}
    \end{subfigure}

\caption{Figure \ref{fig:TheTopic} provides a visualization of the topics used to generate the sample.  Figure \ref{fig:Cheatmap} represents the hierarchy recovered with the proposed method. Table \ref{fig:tab} reports the average and  standard deviation over 10 runs of the clustering accuracy for the various methods, along with average running times.}
\end{figure}

In order to test the ability of SIDIWO to recover latent structures in data, we generate a dataset distributed as a single topic model~(with a vocabulary of 100 words)  whose $8$ topics have an intrinsic hierarchical structure depicted in Figure~\ref{fig:TheTopic}. In this figure,  topics are on the $x$-axis, words on the $y$-axis, and 
 green (resp. red) points represents high~(resp low) probability. 
 We see for example that the first 4 topics are concentrated over the 1st half of the vocabulary, and that topics 1 and 2 have high probability on the 1st and 3rd fourth of the words while for the other two it is on the 1st and 4th.

% In this section we test the ability of the proposed method to recover latent structures in the data. First we generate a dataset distributed as a single topic model whose topics have an intrinsic hierarchical structure: we assume to have a vocabulary of 100 words and 8 topics: the first 4 topics, have a very poor probability for the last 50 words of the vocabulary, while the last 4 topics on the first 50. Focusing on the first 4 topics, the first two have high probabilities on the first and on the third fourth of the first 50 words, while the other two, on the first and on the fourth; the same holds symmetrically for the other 4 topics. To have a visual idea of the topics, we draw them in Figure \ref{fig:TheTopic}: the $x$-axis represents the topics, while $y$-axis contains the words. Red points indicate low probabilities, clear green points higher probabilities.

We generate $400$ samples according to this model and we iteratively run Algorithm~\ref{alg:2} to create a hierarchical binary tree with 8 leafs. We expect leafs to contain samples from a unique topic and internal nodes to gather similar topics. Results are displayed in Figure~\ref{fig:Cheatmap} where each chart represents a node of the tree~(child nodes lay below their parent) and contains the heatmap of the samples clustered in that node~($x$-axis corresponds to samples and $y$-axis to words, red points are  infrequent words and clear points frequent ones). The results are as expected: each leaf contains samples from one of the topics and internal nodes group similar topics together.

% We generate $400$ samples according to this model and we iteratively run Algorithm~\ref{alg:2} to create a hierarchical binary tree with 8 leafs. Ideally, we expect each leaf to contain samples of a unique topic, and the nodes to contain mixture of topics that are similar, according to the hierarchical structure. The results of the hierarchical clustering task is displayed in Figure \ref{fig:Cheatmap}; each chart of the Figure represents a node of the tree; child nodes lay below their parent. Each node represents the heatmap of the samples contained in that node. $x$-axis are the various samples, while $y$-axis are the various words; again, red points are the infrequent words, clear points, the frequent ones. If we focus on the leafs, we can see that each one of them present a pattern that is identical to one of the topics of Figure \ref{fig:TheTopic}; furthermore, the intermediate nodes group together similar topics, providing in this way the expected behavior.

We compare the clustering accuracy of SIDIWO with other methods  using the Adjusted Rand Index \cite{hubert1985comparing} of the partition of the data obtained at the leafs w.r.t the one obtained using the true topics; comparisons are with the flat clustering on $k=8$ topics with TPM, the method from~\cite{SpectralLatent} (SVD), the one from~\cite{kuleshov2015tensor} (Rand. Proj.) and ALS from \cite{kolda2009tensor}, where ALS is applied to decompose a whitened $8\times 8\times 8$ tensor $T$, calculated as in Equation \eqref{eq:TPMT}. We repeat the experiment 10 times with different random samples and we report the average results in Table~\ref{fig:tab}; SIDIWO always recovers the original topic almost perfectly, unlike competing methods. One intuition for this improvement is that each split in the divisive clustering helps remove noise in the moments.

% We analyze the clustering accuracy of the proposed method (named ''SIDIWO'' in the table), comparing the partition of the data obtained on the leafs with the true topic to which each text belongs, using Adjusted Rand Index \cite{hubert1985comparing}, and we compare it with the clustering accuracy obtained performing flat clustering on $k=8$ topics with TPM, with the the method from \cite{SpectralLatent} (named ''SVD'') and from the method in \cite{kuleshov2015tensor} (named ''Rand. Proj.''). The results are displayed in Table \ref{fig:tab}.  We can see that the proposed method perfectly recovers the original topic, unlike competing methods. The intuition of this fact is that the proposed divisive clustering method helps to separate different data; at each step the set of data to be clustered will be less noisy, providing in this way improved performances.

%%%%%%%%%%%%%%%%%%%%%%%%%%%%%%%%%%%%%%%%%%%%%%%%%%%%%%%%%
%\begin{table}[t]
%  %\caption{The clustering performance of the proposed method, compared with flat clustering with competing methods.}
%  \label{tab:clerr}
%  \centering
%  \begin{tabular}{ l  c c c}
%  \toprule
%   \multicolumn{1}{c}{} & \multicolumn{2}{c}{\text{Adj. Rand Idx}} & \multicolumn{1}{c}{} \\ 
%   \text{{Method}}&\text{{Mean}}&\text{{St. dev.}}&\text{{Run. Time}} \\\midrule
%\midrule
%ALS & 0.70 & 0.17 & 1.9 sec.\\\midrule
%TPM & 0.93 & 0.06 & 1.2 sec.\\\midrule
%SVD & 0.52 & 0.13 & 0.1 sec.\\\midrule
%Rand. Proj. & 0.72 & 0.06  & \text{16 min.}\\\midrule
%SIDIWO & 0.98  & 0.01 & 0.4 sec.\\\midrule
%\bottomrule
%\end{tabular}
% \end{table}
%%

\subsection{Experiment on NIPS Conference Papers 1987-2015}\label{sec:NIPS}

\begin{figure}
\centering
\includegraphics[width=\textwidth]{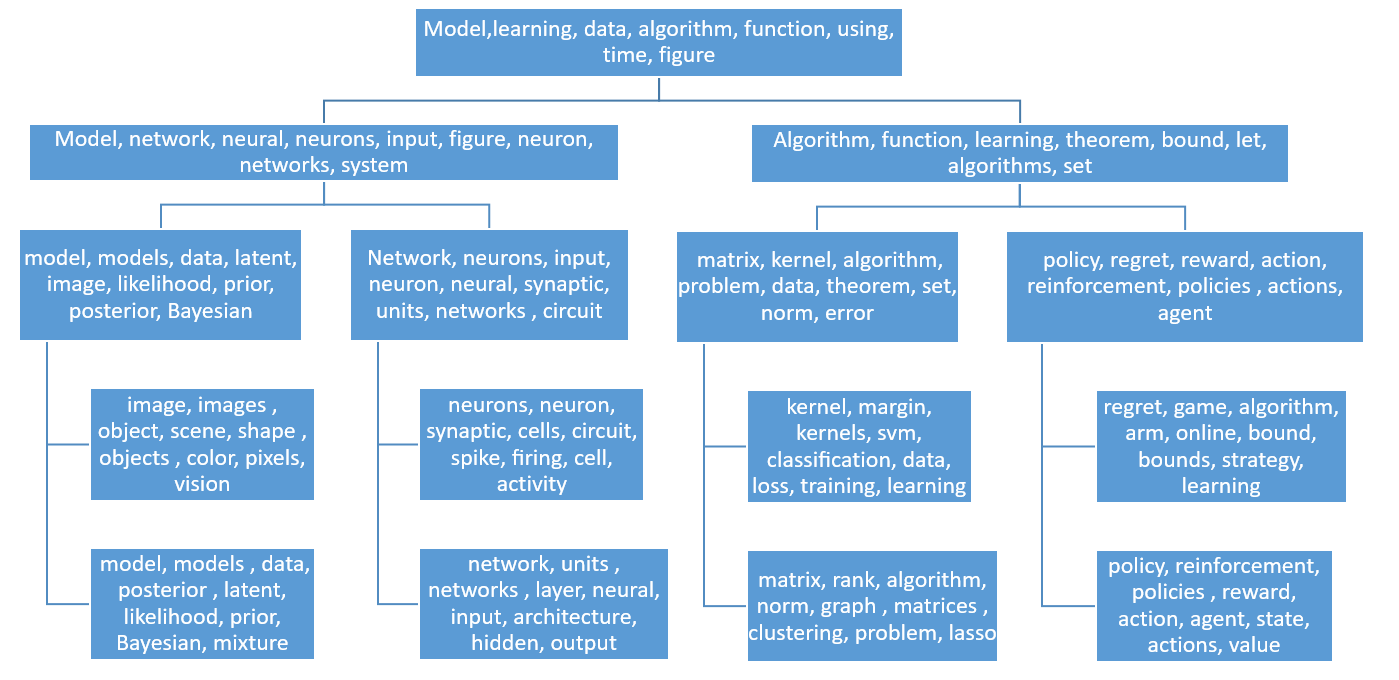}
 \caption{Experiment on the NIPS dataset.}\label{fig:NIPS}
\end{figure}
 
We consider the full set of NIPS papers accepted between 1987 and 2015 , containing $n = 11,463$ papers \cite{perrone2016poisson}. 
We assume that the papers are distributed according to a single topic model, we keep the $d= 3000$ most frequent words as vocabulary and we iteratively run Algorithm~\ref{alg:2} to create a binary tree of depth 4. The resulting tree is shown in Figure~ \ref{fig:NIPS} where each node contains the most \textit{relevant} words of the cluster, where the \textit{relevance}~\cite{sievert2014ldavis} of a word $w\in\mathcal{C}_{node}\subset \mathcal{C}$ is defined by 
$$
 r(w,\mathcal{C}_{node})= \lambda\log\P[w|\mathcal{C}_{node}] +(1-\lambda  )\log \frac{ \P[w|\mathcal{C}_{node}]}{ \P[w|\mathcal{C}]} \enspace,
$$
where the weight parameter is set to $\lambda = 0.7$
and $ \P[w|\mathcal{C}_{node}]$ (resp. $ \P[w|\mathcal{C}]$) is the empirical frequency of $w$ in $\mathcal{C}_{node}$ (resp. in $\mathcal{C}$). 
The leafs clustering and the whole hierarchy have a neat interpretation. Looking at the leaves, we can easily hypothesize the dominant topics for the 8 clusters. From left to right we have: [image processing, probabilistic models], [neuroscience, neural networks], [kernel methods, algorithms], [online optimization, reinforcement learning]. 
Also, each node of the lower levels gathers meaningful keywords, confirming the ability of the method to hierarchically find meaningful topics. The running time for this experiment was 59 seconds.

% We consider the full set of NIPS papers accepted between 1987 and 2015 \footnote{The full dataset can be found here \url{https://archive.ics.uci.edu/ml/datasets/NIPS+Conference+Papers+1987-2015}}, containing $n = 11463$ different papers. We assume the texts of this corpus to be generated by the single topic model, and we iteratively run Algorithm~\ref{alg:2} to create a binary tree, halting when the we reach the fourth level of the tree. For each node of the tree, we then represent, the most \textit{relevant} words, where the \textit{relevance}  \cite{sievert2014ldavis} of a of term $w$ to the subset $\mathcal{C}_{node}\subset \mathcal{C}$ given a weight parameter $\lambda$ is defined as
% $$
%  r(w,\mathcal{C}_{node})= \lambda\log( \P[w|\mathcal{C}_{node}])+(1-\lambda  )\log(\frac{ \P[w|\mathcal{C}_{node}]}{ \P[w|\mathcal{C}]}) ,
% $$
% where $ \P[w|\mathcal{C}_{node}]$ (resp. $ \P[w|\mathcal{C}]$) is the empirical probability of getting the word $w$ in $\mathcal{C}_{node}$ (resp. $\mathcal{C}$). 
% The resulting tree is displayed in figure \ref{fig:NIPS} 
% We can immediately notice that both, the leafs and the whole hierarchy make sense. Looking at the leaves, we can immediately hypothesize the dominant topics of each of the 8 created clusters. From left to right we have: image processing, neural nets, topic modeling, Bayesian statistics, clustering, SVM, reinforcement learning and theory. 
% Looking at the lower levels we can see that at each node we get meaningful keywords, confirming the ability of the method to find meaningful topics in a hierarchical way.

\subsection{Experiment on Wikipedia Mathematics Pages}
\begin{figure}
\centering
\includegraphics[width=\textwidth]{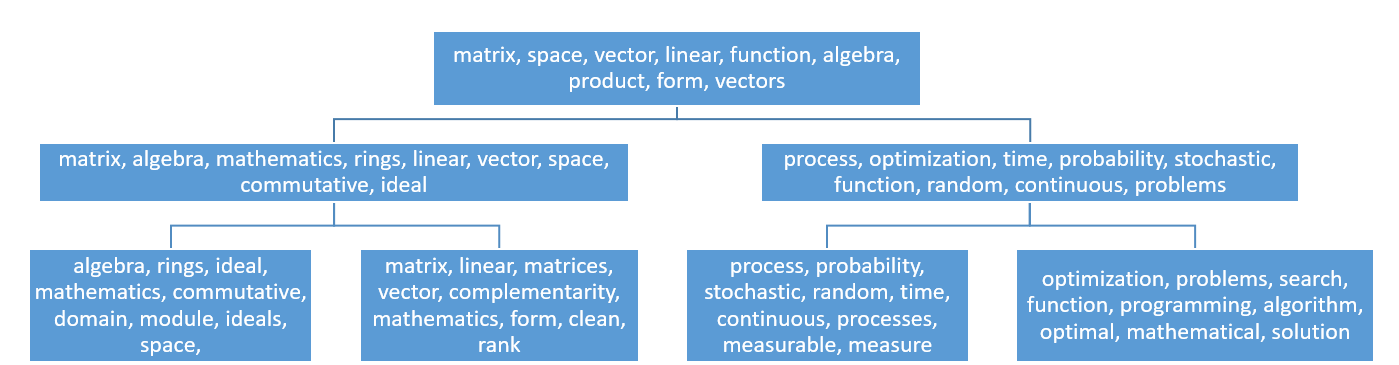}
 \caption{Experiment on the Wikipedia Mathematics Pages dataset.}\label{fig:WIKI}
\end{figure}
We consider a subset of the full Wikipedia corpus, containing all  articles~($n=809$ texts) from the following math-related categories: linear algebra, ring theory, stochastic processes and optimization. 
We remove a set of $895$ stop-words, keep a vocabulary of $d=3000$ words and run SIDIWO to perform hierarchical topic modeling~(using the same methodology as in the previous section). 
The resulting hierarchical clustering is shown in Figure~\ref{fig:WIKI} where we see that each leaf is characterized by one of the dominant topics: [ring theory, linear algebra], [stochastic processes, optimization]~(from left to right). It is interesting to observe that the first level of the clustering has separated pure mathematical topics from applied ones. The running time for this experiment was 6 seconds.

% \subsubsection{Clustering Wikipedia Mathematic Pages}
% \begin{figure}
% \centering
% \includegraphics[width=\textwidth]{WikiSmall.png}
%  \caption{A test on the Wikipedia dataset.}\label{fig:WIKI}
% \end{figure}
% We consider a subset of the full Wikipedia corpus, consisting in all the articles of the following categories: linear algebra, ring theory, stochastic processes and optimization. The total size of the is  of $n=809$ texts, all of them dealing with one main subject: math. After removing a set of $895$ stop-words, we run on the dataset the proposed algorithm to perform hierarchical topic modeling, repeating the same experiment of Section \ref{sec:NIPS}. 
% As in the previous experiment, a binary tree is generated by subsequent binary partitions, and the most relevant words are then represented in each node. Results are plotted in Figure \ref{fig:WIKI}. If we look at the most relevant words of the leafs, we can see that each leaf is characterized by a one dominant topic; from left to right we have: Ring theory, linear Algebra, stochastic processes and optimization. It is interesting to notice that the Algorithm has automatically separated in the first step the pure mathematical topics (on the center-left node) from the applied ones.

\section{Conclusions and future works}
We proposed a novel spectral algorithm~(SIDIWO) that generalizes recent method of moments algorithms relying on tensor decomposition. While previous algorithms lack robustness to model misspecification,  SIDIWO provides meaningful results even in misspecified settings. Moreover, SIDIWO can be used to perform hierarchical method of moments estimation for latent variable models. In particular, we showed through hierarchical topic modeling experiments on synthetic and real data that SIDIWO provides meaningful results while being very computationally efficient.  

A natural future work is to investigate the capability of the proposed hierarchical method to learn overcomplete latent variable models, a task that has received significant attention in recent literature \cite{anandkumar2015learning,anandkumar2017analyzing}. We are also interested in comparing the learning performance of SIDIWO  the with those of other existing methods of moments in the realizable setting. 
On the applications side, we are interested in applying the methods developed in this paper to the healthcare analytics field, for instance to perform hierarchical clustering of patients using electronic healthcare records or more complex genetic data.
\vskip 0.2in
{
\subsubsection*{Acknowledgments}
Guillaume Rabusseau acknowledges support of an IVADO postdoctoral fellowship. Borja Balle completed this work while at Lancaster University.
\small
\bibliography{AR_biblio} 

\begin{thebibliography}{10}

\bibitem{EMDempster}
Arthur~P. Dempster, Nan~M. Laird, and Donald~B. Rubin.
\newblock Maximum likelihood from incomplete data via the {EM} algorithm.
\newblock {\em Journal of the Royal Statistical Society. Series B
  (methodological)}, pages 1--38, 1977.

\bibitem{TensorLatent}
Animashree Anandkumar, Rong Ge, Daniel Hsu, Sham~M Kakade, and Matus Telgarsky.
\newblock Tensor decompositions for learning latent variable models.
\newblock {\em Journal of Machine Learning Research}, 15(1):2773--2832, 2014.

\bibitem{SpectralLatent}
Animashree Anandkumar, Daniel Hsu, and Sham~M Kakade.
\newblock A method of moments for mixture models and hidden {M}arkov models.
\newblock In {\em COLT}, volume~1, page~4, 2012.

\bibitem{SpectralLDA}
Animashree Anandkumar, Yi-kai Liu, Daniel~J Hsu, Dean~P Foster, and Sham~M
  Kakade.
\newblock A spectral algorithm for {L}atent {D}irichlet {A}llocation.
\newblock In {\em NIPS}, pages 917--925, 2012.

\bibitem{jain2014learning}
Prateek Jain and Sewoong Oh.
\newblock Learning mixtures of discrete product distributions using spectral
  decompositions.
\newblock In {\em COLT}, pages 824--856, 2014.

\bibitem{hsu2013learning}
Daniel Hsu and Sham~M Kakade.
\newblock Learning mixtures of spherical {G}aussians: moment methods and
  spectral decompositions.
\newblock In {\em ITCS}, pages 11--20. ACM, 2013.

\bibitem{song2011spectral}
Le~Song, Eric~P Xing, and Ankur~P Parikh.
\newblock A spectral algorithm for latent tree graphical models.
\newblock In {\em ICML}, pages 1065--1072, 2011.

\bibitem{EMSlow}
Borja Balle, William~L Hamilton, and Joelle Pineau.
\newblock Methods of moments for learning stochastic languages: Unified
  presentation and empirical comparison.
\newblock In {\em ICML}, pages 1386--1394, 2014.

\bibitem{chaganty2013spectral}
Arun~T Chaganty and Percy Liang.
\newblock Spectral experts for estimating mixtures of linear regressions.
\newblock In {\em ICML}, pages 1040--1048, 2013.

\bibitem{bailly2011quadratic}
Raphael Bailly.
\newblock Quadratic weighted automata: Spectral algorithm and likelihood
  maximization.
\newblock {\em Journal of Machine Learning Research}, 20:147--162, 2011.

\bibitem{zhang2014spectral}
Yuchen Zhang, Xi~Chen, Denny Zhou, and Michael~I Jordan.
\newblock Spectral methods meet {EM}: A provably optimal algorithm for
  crowdsourcing.
\newblock In {\em NIPS}, pages 1260--1268, 2014.

\bibitem{steinbach2000comparison}
Michael Steinbach, George Karypis, Vipin Kumar, et~al.
\newblock A comparison of document clustering techniques.
\newblock In {\em KDD workshop on text mining}, volume 400, pages 525--526.
  Boston, 2000.

\bibitem{savaresi2001performance}
Sergio~M Savaresi and Daniel~L Boley.
\newblock On the performance of bisecting {K}-means and {PDDP}.
\newblock In {\em SDM}, pages 1--14. SIAM, 2001.

\bibitem{DBLP:conf/icml/BalleQC12}
Borja Balle, Ariadna Quattoni, and Xavier Carreras.
\newblock Local loss optimization in operator models: a new insight into
  spectral learning.
\newblock In {\em ICML}, pages 1819--1826, 2012.

\bibitem{DBLP:conf/nips/BalleM12}
Borja Balle and Mehryar Mohri.
\newblock Spectral learning of general weighted automata via constrained matrix
  completion.
\newblock In {\em NIPS}, pages 2159--2167, 2012.

\bibitem{DBLP:conf/icml/QuattoniBCG14}
Ariadna Quattoni, Borja Balle, Xavier Carreras, and Amir Globerson.
\newblock Spectral regularization for max-margin sequence tagging.
\newblock In {\em ICML}, pages 1710--1718, 2014.

\bibitem{kulesza2014low}
Alex Kulesza, N~Raj Rao, and Satinder Singh.
\newblock Low-rank spectral learning.
\newblock In {\em Artificial Intelligence and Statistics}, pages 522--530,
  2014.

\bibitem{kulesza2015low}
Alex Kulesza, Nan Jiang, and Satinder Singh.
\newblock Low-rank spectral learning with weighted loss functions.
\newblock In {\em Artificial Intelligence and Statistics}, pages 517--525,
  2015.

\bibitem{ruffini2016new}
Matteo Ruffini, Marta Casanellas, and Ricard Gavald{\`a}.
\newblock A new spectral method for latent variable models.
\newblock {\em arXiv preprint arXiv:1612.03409}, 2016.

\bibitem{SpectralLatentHMM}
Daniel Hsu, Sham~M Kakade, and Tong Zhang.
\newblock A spectral algorithm for learning hidden {M}arkov models.
\newblock {\em Journal of Computer and System Sciences}, 78(5):1460--1480,
  2012.

\bibitem{kolda2009tensor}
Tamara~G Kolda and Brett~W Bader.
\newblock Tensor decompositions and applications.
\newblock {\em SIAM review}, 51(3):455--500, 2009.

\bibitem{kuleshov2015tensor}
Volodymyr Kuleshov, Arun Chaganty, and Percy Liang.
\newblock Tensor factorization via matrix factorization.
\newblock In {\em AISTATS}, pages 507--516, 2015.

\bibitem{cardoso1996jacobi}
Jean-Francois Cardoso and Antoine Souloumiac.
\newblock Jacobi angles for simultaneous diagonalization.
\newblock {\em SIAM journal on matrix analysis and applications},
  17(1):161--164, 1996.

\bibitem{bunse1993numerical}
Angelika Bunse-Gerstner, Ralph Byers, and Volker Mehrmann.
\newblock Numerical methods for simultaneous diagonalization.
\newblock {\em SIAM journal on matrix analysis and applications},
  14(4):927--949, 1993.

\bibitem{duchi2008efficient}
John Duchi, Shai Shalev-Shwartz, Yoram Singer, and Tushar Chandra.
\newblock Efficient projections onto the l 1-ball for learning in high
  dimensions.
\newblock In {\em ICML}, pages 272--279, 2008.

\bibitem{van2011numpy}
Stefan Van Der~Walt, S~Chris Colbert, and Gael Varoquaux.
\newblock The numpy array: a structure for efficient numerical computation.
\newblock {\em Computing in Science \& Engineering}, 13(2):22--30, 2011.

\bibitem{hubert1985comparing}
Lawrence Hubert and Phipps Arabie.
\newblock Comparing partitions.
\newblock {\em Journal of classification}, 2(1):193--218, 1985.

\bibitem{perrone2016poisson}
Valerio Perrone, Paul~A Jenkins, Dario Spano, and Yee~Whye Teh.
\newblock Poisson random fields for dynamic feature models.
\newblock {\em arXiv preprint arXiv:1611.07460}, 2016.

\bibitem{sievert2014ldavis}
Carson Sievert and Kenneth~E Shirley.
\newblock Ldavis: A method for visualizing and interpreting topics.
\newblock In {\em ACL workshop on interactive language learning, visualization,
  and interfaces}, 2014.

\bibitem{anandkumar2015learning}
Animashree Anandkumar, Rong Ge, and Majid Janzamin.
\newblock Learning overcomplete latent variable models through tensor methods.
\newblock In {\em COLT}, pages 36--112, 2015.

\bibitem{anandkumar2017analyzing}
Animashree Anandkumar, Rong Ge, and Majid Janzamin.
\newblock Analyzing tensor power method dynamics in overcomplete regime.
\newblock {\em Journal of Machine Learning Research}, 18(22):1--40, 2017.

\bibitem{robeva2016decomposing}
Elina~Mihaylova Robeva.
\newblock {\em Decomposing Matrices, Tensors, and Images}.
\newblock PhD thesis, University of California, Berkeley, 2016.

\end{thebibliography}
\bibliographystyle{unsrt}
}

\newpage
\appendix
{
\section{Low-rank whitenings may not admit a symmetric orthogonal decomposition.}
In Section \ref{subsec:misspecified} we claimed that a  symmetric tensor with CP-rank $k$, when whitened to a $l\times l\times l$ tensor $T_l$, may not admit a symmetric orthogonal decomposition if $l<k$. We give here a simple counter-example by constructing a tensor $M_3\in\R^{3\times 3\times 3}$ whose $2\times 2\times 2$ whitening does not admit a symmetric orthogonal decomposition. We will make use of the following Lemma.
\begin{lemma}[\cite{robeva2016decomposing}, Example 1.2.3]\label{lemma:odeco}
A $2\times 2 \times 2$ symmetric tensor $T$ is orthogonally decomposable if and only if its entries satisfy the following equation:
\begin{equation}\label{eq:odeco}
(T)_{1,1,1}(T)_{2,2,1} + (T)_{2,1,1}(T)_{2,2,2} = (T)_{2,1,1}^2 + (T)_{2,2,1}^2.
\end{equation}
\end{lemma}
Consider the following parameters\footnote{This example easily generalizes to vectors in the simplex.}
$$
\mu_1 = \bmat{1\\0\\0},\ \ \mu_2 = \bmat{0\\1\\0},\ \ \mu_3 = \bmat{1\\1\\1},\ \omega = \bmat{1\\1\\1}
$$
from which one can recover a matrix $M_2$ and a tensor $M_3$ from equations \eqref{eqn:mom2} and \eqref{eqn:mom3}, both of rank $k=3$. 
Using the top $2$ singular vectors and values of $M_2$, $M_3$ would be whitened  to a $2\times 2 \times 2$ tensor $T$ with the following entries:
%\begin{eqnarray*}
%&(T)_{1,1,1} = 2a^3+c^3,\\ 
%&(T)_{1,2,2} = (T)_{2,1,2} = (T)_{2,2,1} = 2ab^2,\\ 
%&(T)_{1,2,1} = (T)_{1,1,2} = (T)_{2,2,2} = (T)_{2,1,1} = 0
%\end{eqnarray*}
%
% 
%where 
%$$
%a=\frac{z}{n\sqrt{1+2z}},\ \ b = \frac{1}{\sqrt{2}},\ \ c = \frac{2z+1}{n\sqrt{1+2z}}
%$$
%and
%$$
%n = \sqrt{2z^2+1},\ \ z = \frac{1+\sqrt{3}}{2}.
%$$
\begin{eqnarray*}
&(T)_{1,1,1} = 2(\frac{1+\sqrt{3}}{2\sqrt{9+5\sqrt{3}}})^3+(\frac{2+\sqrt{3}}{\sqrt{9+5\sqrt{3}}})^3,\\ 
&(T)_{1,2,2} = (T)_{2,1,2} = (T)_{2,2,1} = \frac{1+\sqrt{3}}{2\sqrt{9+5\sqrt{3}}},\\ 
&(T)_{1,2,1} = (T)_{1,1,2} = (T)_{2,2,2} = (T)_{2,1,1} = 0
\end{eqnarray*}
One can check that in this case Eq.~\eqref{eq:odeco} is equivalent to $(T)_{1,1,1}= (T)_{2,2,1}$, which does not hold, hence $T$ is not orthogonally decomposable. } 
%We claim that $T_2$ is not orthogonally decomposable. To see this, use the result in Lemma \ref{lemma:odeco}, and observe that the condition in equation \eqref{eq:odeco} is not satisfied.}

\section{Proof Of Lemma \ref{lemma:1}}
Consider the SVD of $M_2$:
$$
M_2 = U S U^\top
$$
where $U \in \R^{d\times k}$ and $S \in \R^{k\times k}$ are obtained from the first $k$ singular vectors and values. Define now the matrix $E =  U S^{1/2}$; then there exists a unique $k\times k$ orthogonal matrix $O$ such that $M\Omega^{1/2}  = E O$.

This implies that the slices of $M_3$ can be rewritten as follows:
$$
M_{3,r} =   M \,\Omega^{1/2}  diag(m_r) \,\Omega^{1/2} \, M^\top =  E O diag(m_r) \,(E O)^\top.
$$
Take now a generic matrix $D \in \mathcal{D}_k$; it can be written as
$$
D = O_k^\top E^\dagger
$$
for a certain orthonormal matrix $O_k$. So we have
$$
D M_{3,r}D^\top = O_k^\top E ^\dagger E O diag(m_r) \,(E O )^\top (O_k^\top E ^\dagger)^\top = O_k^\top  O diag(m_r) \, O O_k  
$$
This matrix is diagonal if and only if $ O = O_k$, so the problem
\begin{equation}
\min_{D\in \mathcal{D}_k} \sum_{i\neq j}(\sum_{r=1}^d (D M_{3,r}D^\top)^2_{i,j} )^{1/2}
\end{equation}
is optimized by $D = (M\Omega^{1/2})^\dagger = (E O)^\dagger$, which is the unique (up to a columns rescaling) feasible optimum.   

\section{Proof Of Theorem \ref{prop:whiten}}
Let's recall the notation we are going to use. 
Consider the matrix $M_2$ and its SVD:
$$
M_2 = USU^\top
$$
For any $l\leq k$, define $E_l = U_lS_l^{1/2}  \in \R^{d\times l}$, where $U_l$ and $S_l$ are $U$ and $S$ truncated at the $l$-th singular vector (recall: $U \in \R^{d\times k}$ and $S \in \R^{k\times k}$). We know that there exists an orthonormal matrix $O$ such that  
$$
M\Omega^{1/2} = E O
$$
Let's prove the various points of the Theorem.
\begin{enumerate}
\item Consider any matrix $D\in\mathcal{D}_l$. Then we will have, for an orthonormal $O_l$,
$$
D = (E_lO_l)^\dagger = O_l^\top S_l^{-1/2}U_l^\top
$$
To prove the statement, it is enough to show that the matrix 
$
C = DM\Omega^{1/2}
$ has rank $l$. To see this, explicitly represent $C$:
$$
C = DM\Omega^{1/2} = O_l^\top S_l^{-{1/2}}U_l^\top E O = O_l^\top S_l^{-{1/2}}U_l^\top U S^{1/2} O = O_l^\top\mathbb{I}_{l,k}O
$$
the fact that $O$ and $O_l$ are orthogonal proves the claim.
\item Consider again any matrix $D\in\mathcal{D}_l$, then 
$$ 
D^\dagger = {\tilde{M}} {\tilde{\Omega}}^{1/2} = (E_lO_l) = U_lS_l^{1/2}O_l
$$
The columns of $U_l$ are the left singular vectors of $M\Omega^{1/2}$, that span the best fit $l$-dimensional subspace of the space generated by the columns of $M$.
\item To prove this we will proceed by contradiction.
Assume that $(M\Omega^{1/2}\mathbb{I}_{k,l})^\dagger \in \mathcal{D}_l$; this means that there exists an orthonormal matrix $O_l$ such that
$$
M\Omega^{1/2}\mathbb{I}_{k,l} = E_lO_l = E\mathbb{I}_{k,l}O_l
$$
But $M\Omega^{1/2} = E O$, so
$$
E O\mathbb{I}_{k,l}  = E\mathbb{I}_{k,l}O_l
$$
This would imply that 
$$
\mathbb{I}_{l,k}O\mathbb{I}_{k,l}  = O_l
$$
and so, for some $P\in \R^{k-l\times k-l}$
$$
O = \left[ \begin{array}{c|c}
O_l & 0\\ \hline
0 & P
\end{array}\right]
$$
Observe now that the matrix $Z = \Omega^{1/2}M^\top M\Omega^{1/2} $ has all the entries that are different from zero, by the hypothesis that $\langle \mu_i, \mu_j \rangle \neq 0$ for any $i,j$. However, we have that 
$$
Z = \Omega^{1/2}M^\top M\Omega^{1/2} = O^\top S O = 
$$
$$
= \left[ \begin{array}{c|c}
O_l^\top & 0\\ \hline
0 & P^\top
\end{array}\right] \left[ \begin{array}{c|c}
S_l & 0\\ \hline
0 & S_{l,k}
\end{array}\right]  \left[ \begin{array}{c|c}
O_l & 0\\ \hline
0 & P
\end{array}\right] = \left[ \begin{array}{c|c}
O_l^\top S_l O_l& 0\\ \hline
0 & P^\top S_{l,k} P
\end{array}\right]  
$$
where $S_{l,k}$ is the diagonal matrix with the last $k-l$ singular values. So $Z$ has some zero entry. This contradiction proves the claim.

The proof of the fact that $ \mathbb{I}_{l,k}(M_\pi\Omega_\pi^{1/2})^\dagger\notin \mathcal{D}_l$ is identical.
\end{enumerate}

\section{Proof Of Theorem \ref{teo:newopt}}
Recall the considered problem:
\begin{equation}
\min_{D\in\mathcal{D}_l}  \sum_{i\neq j} \sup_{v\in \mathcal{V}_M} \sum_{h = 1}^{k}\langle d_i,\mu_h\rangle\langle d_j,\mu_h\rangle\omega_h v_h 
\end{equation}
where
$$
\mathcal{V}_M = \{v\in\R^k: v = \alpha^\top M,  \text{where } \|\alpha\|_2 \leq 1\}
$$

Consider any $v\in\mathcal{V}_M$, then it admits the following representation, for some $\alpha$ with $\|\alpha\|_2 \leq 1$:
$$
v = [\langle \alpha,\mu_1 \rangle,...,\langle \alpha,\mu_k\rangle]^\top.
$$
This allows the following chain of equalities on the cost function:
\begin{align*}
\sum_{i\neq j} \sup_{v\in \mathcal{V}_M} \sum_{h = 1}^{k}\langle d_i,\mu_h\rangle\langle d_j,\mu_h\rangle\omega_h v_h 
&= 
 \sum_{i\neq j} \sup_{\alpha:\|\alpha \|_2\leq 1} \sum_{h = 1}^{k}\langle \alpha,\mu_h\rangle \langle d_i,\mu_h\rangle \langle d_j,\mu_h\rangle \omega_h  \\
&= 
\sum_{i\neq j} \sup_{\alpha:\|\alpha \|_2\leq 1}  \sum_{r = 1}^{d} \alpha_r\sum_{h = 1}^{k}\mu_{h,r}\langle d_i,\mu_h\rangle \langle d_j,\mu_h\rangle \omega_h  \\
&= 
\sum_{i\neq j} \sup_{\alpha:\|\alpha \|_2\leq 1} \sum_{r=1}^{d}\alpha_r( D M_{3,r}D^\top)_{i,j}  \\
&=
\sum_{i\neq j} \sup_{\alpha:\|\alpha \|_2\leq 1}\langle \alpha,t_{i,j}\rangle  \\
&=  
\sum_{i\neq j}\|t_{i,j}\|_2
\end{align*}
Where the vector $t_{i,j}$ is defined as 
$$
t_{i,j} = ((D M_{3,1}D^\top)_{i,j},...,( D M_{3,d}D^\top)_{i,j})
$$
and the last equality has been obtained from the fact that, for any vector $w \in \R^k$, we have
$$
\|w\| = \sup_{\alpha:\|\alpha \|_2\leq 1}\langle \alpha,w \rangle  
$$
This last equation proves our statement; in fact, 
$$
 \sum_{i\neq j}\|t_{i,j}\|_2 =  \sum_{i\neq j}(\sum_{r=1}^d (D M_{3,r}D^\top)^2_{i,j})^{1/2}.
$$

\section{Proof Of Theorem \ref{teo:l2}}
First, observe that the set of $2\times 2$ orthonormal matrices can be parametrized as
\begin{equation}\label{eq:Ol}
O_a = 
  \begin{bmatrix}
    \sqrt{1-a^2} &a\\
    -a & \sqrt{1-a^2}
  \end{bmatrix},\,\,\,\text{for  }a \in [-1,1].
\end{equation} 
The set $\mathcal{D}_2$ can thus be rewritten in function of $a$, as
$$
\mathcal{D}_2 = \{D: D = (E_2 O_a)^\dagger \text{ for  }a \in [-1,1]\}.
$$ 
and Problem~\eqref{eq:constropt} can be rewritten as
$$
\min_{a \in [-1,1]}\sum_{r=1}^d 2(O_a^\top H_{2,r}O_a)^2_{1,2}
$$
where 
 \begin{equation} 
H_{2,r} = E_2^{\dagger} M_{3,r}E_2^{\dagger\top},\,\,\, for \,\,r=\{1,...,d\}
\end{equation}
and where we used the fact that
$
O_a^\top H_{2,r}O_a
$
 is symmetric. 
 We can then write
 $$
 (O_a^\top H_{2,r}O_a)^2_{1,2} = c_1^{(r)} a^4 + c_2^{(r)} a^3\sqrt{1-a^2} + c_3^{(r)} a\sqrt{1-a^2} + c_4^{(r)} a^2 + c_5^{(r)}
 $$
 where the coefficients can be written as
$$
c_1^{(r)} = 4h^2 - f^2, \ \ \ \  c_2^{(r)} = -4fh,  \ \ \ \  c_3^{(r)} =  2fh, \ \ \ \ 
c_4^{(r)} = f^2 - 4h^2,\ \ \ \ c_5^{(r)} = h^2 
$$
with 
$h = (H_{2,r})_{1,2}$ and $f = (H_{2,r})_{1,1} - (H_{2,r})_{2,2}$.
Letting 
$
c_j = \sum_{r=1}^d c_j^{(r)}
$
for  $j \in \{1,...,5\}$
it follows that optimizing Problem~\eqref{eq:constropt} is equivalent to minimize the following smooth real function
$$
 F(a) = c_1 a^4 + c_2 a^3\sqrt{1-a^2} + c_3 a\sqrt{1-a^2} + c_4  a^2 + c_5.
 $$
\end{document}